\documentclass[sn-mathphys-num]{sn-jnl}

\usepackage{graphicx}%
\usepackage{multirow}%
\usepackage{amsmath,amssymb,amsfonts}%
\usepackage{amsthm}%
\usepackage{mathrsfs}%
\usepackage[title]{appendix}%
\usepackage{xcolor}%
\usepackage{textcomp}%
\usepackage{manyfoot}%
\usepackage{booktabs}%
\usepackage{algorithm}%
\usepackage{algorithmicx}%
\usepackage{algpseudocode}%
\usepackage{listings}%

\theoremstyle{thmstyleone}%
\theoremstyle{thmstyletwo}%

\theoremstyle{thmstylethree}%

\graphicspath{{figures/}}
\usepackage{subfig}
\begin{document}

\title[Generic Multimodal Spatially Graph Network for Spatially Embedded Network Representation Learning]{Generic Multimodal Spatially Graph Network for Spatially Embedded Network Representation Learning}


\author[1]{\fnm{Xudong} \sur{Fan}}\email{xudongfa@buffalo.edu}

\author*[2]{\fnm{J\"urgen} \sur{Hackl}}\email{hackl@princeton.edu}
\equalcont{These authors contributed equally to this work.}

\affil[1]{\orgdiv{Department of Civil, Structural and Environmental Engineering}, \orgname{SUNY at Buffalo}, \orgaddress{\street{212 Ketter Hall}, \city{Buffalo}, \postcode{14260}, \state{NY}, \country{USA}}}

\affil[2]{\orgdiv{Complex Infrastructure Systems Group}, \orgname{Princeton University}, \orgaddress{\street{54 Olden Street}, \city{Princeton}, \postcode{08544}, \state{NJ}, \country{USA}}}


\abstract{
  Spatially embedded networks (SENs) represent a special type of
  complex graph, whose topologies are constrained by the networks' embedded
  spatial environments. The graph representation of such networks is thereby
  influenced by the embedded spatial features of both nodes and edges. Accurate
  network representation of the graph structure and graph features is a
  fundamental task for various graph-related tasks.  In this study, a Generic
  Multimodal Spatially Graph Convolutional Network (GMu-SGCN) is developed for
  efficient representation of spatially embedded networks. The developed
  GMu-SGCN model has the ability to learn the node connection pattern via
  multimodal node and edge features.  In order to evaluate the developed model,
  a river network dataset and a power network dataset have been used as test
  beds. The river network represents the naturally developed SENs, whereas the
  power network represents a man-made network. Both types of networks are
  heavily constrained by the spatial environments and uncertainties from nature.
  Comprehensive evaluation analysis shows the developed GMu-SGCN can improve
  accuracy of the edge existence prediction task by 37.1\% compared to a
  GraphSAGE model which only considers the node's position feature in a power
  network test bed. Our model demonstrates the importance of considering the
  multidimensional spatial feature for spatially embedded network
  representation.
}




\maketitle

\section{Introduction}\label{sec1}

Many real-world networked infrastructure or natural systems can be represented
as spatially embedded networks (SENs), as their structures are constrained and
shaped by the spatial environments \cite{Nocera2022Selection}. For instance,
road networks can be represented by using intersections as nodes and road
segments as edges \cite{Xu2022Evaluation}, and their topologies are shaped by
land types and socio-economic factors. For another instance, the urban sensor
networks can also be represented by using the sensors as nodes and the sensor
connections as edges \cite{Wang2023STGIN}. The functionalities and status of
such SENs directly influence the city's sustainability and social equity
\cite{Zanfei2022Novel,Li2024Passenger,Li2024Predicting}. To better manage these
networked systems, such as status monitoring and automatic control, advanced
techniques from Artificial Intelligence (AI) and Machine Learning (ML) have
gained significant attention
\cite{lee_urban-net_2016,jiang_graph_2022}. However, the performance of AI and
ML models heavily depends on accurate modeling and robust learning capabilities
for these networked systems. To further improve the networked systems'
management, there is an increasing demand for models that can effectively
capture and represent the complex and intricate patterns of the SENs.

The graph representation learning aims to extract the complex intricate patterns
and relationships within the networks by embedding high-dimensional sparse graph
structured data into low-dimensional dense vectors
\cite{Ju2024Comprehensive}. Various techniques have been developed for efficient
representation learning, including traditional graph embedding methods and
geometric-based deep learning methods. Traditional graph embedding methods can
be broadly grouped into matrix-based methods, random walk-based methods, and
non-Graph Neural Network methods. Specifically, the matrix-based methods
describes the networks with a proximity measure matrix, and the representation
is computed based on that matrix. For example, the locally linear embedding
(LLE) first constructs a weight matrix which includes the linear combination of
node features. The low-dimensional representation for nodes can be then computed
by solving the matrix's eigenvalues \cite{Roweis2000Nonlinear}. Random
walks-based methods are also emerging in graph representation. The key idea
behind random work is that the nodes should have similar features if they tend
to co-occur in random short paths. For example, DeepWalk
\cite{Perozzi2014DeepWalk} and Node2Vec \cite{Grover2016node2vec} are two common
traditional graph embedding methods. The former approach truncated random walks
within a network and represented the networks by treating these walks as
equivalent sentences. The latter approach used biased random walks to improve
the efficiency of exploring diverse neighborhoods. Lastly, non-graph neural
network-based algorithm can also work with graph structure data similar to the
matrix-based methods. However, the dimensional reduce method is replaced with
neural network structures, such as the autoencoder structure
\cite{Wang2016Structural}.

Recently, the geometric-based deep learning algorithms have been emerging. The
Graph Convolutional Network (GCN) was first proposed for graphs' semi-supervised
learning tasks \cite{Kipf2017SemiSupervised}, which forms a foundation
architecture in geometric deep-learning models. After that, several variants of
GCN have been proposed for better network representation and pattern discovery,
including the GraphSAGE \cite{Hamilton2017Inductive}, graph attention network
(GAT) \cite{Velickovic2018Graph}, and many others \cite{Zhou2020Graph}. The
neural networks have greatly extended the flexibility of traditional geometric
algorithms by using a large set of linear transformation layers and non-linear
activation functions \cite{Karlaftis2011Statistical}. Previous studies have
shown that the precise feature representation of either the nodes or the graphs
is the key for the accurate prediction or classification of graphs and their
components. For example, the accurate prediction of traffic flow in a road
network needs to embed the spatial-temporal features into each node
\cite{Ali2022Exploiting}.  For another example, the nodes' representative
vectors of a stormwater distribution system have also been learned by combining
the system topology and the spatial environmental factors, such as the rainfall
data and surface runoff. The node's representative vectors are then used for the
prediction of node and pipe status, i.e., the magnitude of junction inflows and
pipe flow rates \cite{Li2024Predicting}.

As a special type of complex networks, the SENs have been used to model many
real-world physical networked systems. Examples of spatially embedded networks
include road networks, water networks, and river networks
\cite{Dong2020networknetworks, Daqing2011Dimension}. For example, studies have
observed that many real-world SENs can be modeled by the 'small world' and
'Erd\H{o}s-R\'enyi' networks \cite{Daqing2011Dimension}. However, learning the
intricate patterns of SENs is still challenging due to the complex interactions
between the SENs and the embedded environments. Both the node properties and
edge properties may influence the topologies of SENs in different ways. In order
to better learn the intricate patterns of the SENs, a model is expected to have
the capability to use all the properties as input and obtain the relationship
automatically. For example, recent studies have found that the classification
accuracy of the atom network can be improved when adding the node's position
feature into the dataset \cite{Danel2020Spatial}. Previous studies have also
highlighted the influence of road steepness on the walking paths
\cite{Daniel2018How}.  However, due to the different data types and structures
of these features, traditional single-modal learning methods are struggling to
learn the representation from multimodal network features simultaneously.

In order to tackle the multi-modal challenging in the representation leading of
complex SENs, this study develops a generic multimodel graph learning framework,
which is named as Generic Multimodal Spatially Graph Convolutional Network
(GMu-SGCN). The developed model can efficiently embed different spatial features
in to the node's latent represent vectors.  Given the flexibility of the
developed GMu-SGCN model, two variants of the developed GMu-SGCN model, i.e.,
the Regional Spatial Graph Convolutional Network (RSGCN) and Edge Spatial Graph
Convolutional Network (ESGCN), are also introduced. The RSGCN model only
considers the network's node related features while ignores features related to
the edge. On the contrary, the ESGCN model only considers the edge related
features. Two different types of SENs are used in this study to evaluate the
performances of considered models. It should be noted that the performance of
the RSGCN model has been introduced in the authors' previous study
\cite{Fan2024Modelinga}. However, the previous study did not consider the edge
features, which is one of the key feature influences the topology of
SENs. Moreover, unlike the previous study which used the same testbed for
training and testing purposes, this study used two different test beds for
training and testing separately. In summary, the contributions of this paper are
listed as follows:

\begin{itemize}
\item This study proposes a Generic Multimodal Spatially Graph Convolutional
  Network, namely GMu-SGCN model, for the representation learning of complex
  SENs. The developed model can process the multimodal nodes' and edges'
  features simultaneously.
\item This study also proposes a network reconstruction-based framework for
  evaluating the network representation performance of different geometric-based
  deep learning models. The framework firstly partition the large-scale SENs
  into subgraphs and then evaluates the model by the edge existence prediction
  accuracy. This framework can significantly reduce the computational memory in
  the training and prediction processes.
\item The developed model is comprehensively evaluated by two real world
  datasets, a power distribution network and a river network. The high accurate
  reconstruction of both networks have demonstrated that the connection patterns
  of both SENs are influenced by their spatial environment and can be learned by
  the developed model.
\end{itemize}

\section{Related work and motivation}

\textbf{Spatially Embedded Networks (SENs) and Network Connection Modeling}. The
spatially embedded network is a special type of complex network whose structure
is constrained by its embedded spatial environments. A simple synthetic
spatially embedded random network can be created by randomly placing the nodes
and creating edges based on the nodes' distances \cite{Barnett2007Spatially}. In
real world, many natural and man-made network systems show constant patterns
\cite{Dong2020networknetworks}. For example, the lengths of infrastructure
systems are often limited by construction costs in the real world
\cite{Cadini2015cascading}. Previous studies have observed that the lengths of
road segments in a transportation network follow a power-law distribution
\cite{Chen2019Correlations}, and the node degree density within a large power
system follows a logarithmic distribution \cite{Soltan2016Generation}. Because
of such constraints, the connection function within such SENs can be modeled
mathematically, such as the small-world networks have been used to model the
power networks \cite{Aksoy2019generative}, the single-parameter controlled
hierarchical planer and spatial networks \cite{Molinero2020model}, the tunable
spanning tree \cite{Dunton2023Generating}, and a combination of relative
neighborhood graphs (RNG), Gabriel Graphs (GG), and Erd\H{o}s-R\'enyi (ER)
random graph \cite{Hackl2019Modelling}.  Although previous studies have
demonstrated the connection functions of SENs follow specific patterns, it is
still challenging to identify accurate and efficient network connection models.

\begin{figure*}[!htb]
  \centering
  \includegraphics[width=0.6\textwidth]{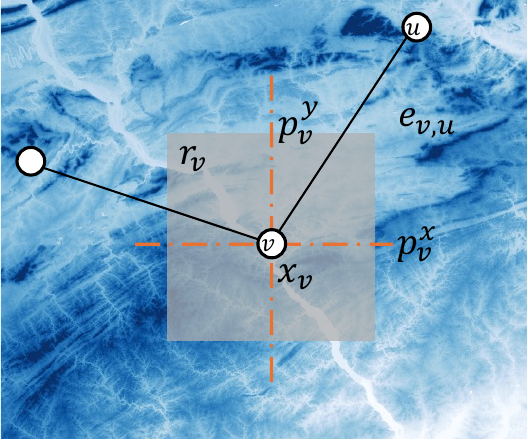}
  \caption{Spatially Embedded Networks ($x_v$ represents the node's point
    feature, $r_v$ represents the node's regional feature, $p_v^y, p_v^x$
    represent the node's position feature, and $e_{v,u}$ represents the edge
    feature between node $u$, and $v$.)}
  \label{fig: spatially embedded networks}
\end{figure*}

To better illustrate the concept of SENs and its relationship with the embedded
spatial environment, Figure \ref*{fig: spatially embedded networks} shows an
example of SEN and its embedded environment. Specifically, four types of
features may influence the connection patterns of the SEN, i.e., the node's
point feature $x$, the node's regional feature $r$, the node's position feature
$pos$, and the edge feature $e$. The node's point feature refers to the spatial
feature that located at the node's point, such as the population density, the
socioeconomic data, and geological data of the node's location. Unlike to the
node's point feature, the node's regional feature describes a region centered on
the node, such as the topography change or soil type change within a specific
distance of a node. The node's position feature is considered separately in this
study, which refers to the node's coordinates values. Lastly, the edge feature
represents a sequence of spatial values that sampled from the spatial
environment along the edges.

\textbf{Deep learning-based network representation learning:} Recently, the deep
learning methods have been emerging and showed more promising performance in
various applications. Unlike conventional mathematically modeled methods which
use rigorous mathematical equations to describe the SENs' intricate patterns,
geometric deep-learning models offer a more flexible and end-to-end learning
process, which facilitates network representation and intricate pattern
discovery \cite{Ding2024Artificial}. The deep learning process of the Graph
Convolutional Networks (GCNs) can be generally described as
Eq. \ref{eq:GCNequation}, i.e. the node features will be processed by the
neurons and then convolved along the edges of the graph. Notable variants of the
GCN include the GraphSAGE \cite{Hamilton2017Inductive} and Spatial Graph
Convolutional Networks (SGCN) \cite{Danel2020Spatial}. The former architecture
introduced advanced sampling strategies for the node's neighbors, resulting in a
higher node classification accuracy in multiple datasets. The latter SGCN
architecture first introduced the spatial features of nodes into the learning
process of a molecular classification task.  The GCN and its variants have been
receiving more and more attention in spatially embedded networks. For example, a
graph attention architecture was used to capture the spatial correlations within
traffic networks for traffic flow prediction \cite{Wang2023STGIN}. The GCNs have
also been used in power systems for fault detection, power outage prediction,
power flow simulation, and system control \cite{Liao2022Review}. However, most
of the GCN variants were mainly focusing on the features of nodes rather than
the edges. Considering the significant influence of both node features and edge
features on the spatially embedded networks, there is an urgent need for models
that can process such heterogeneous features simultaneously.

\begin{equation}
  \label{eq:GCNequation}
  H^{l+1} = \sigma(\hat{A}H^lW^l)
\end{equation}

\textbf{Multimodal Data Fusion:} Multimodal data fusion represents a fundamental
method for mining richer information from data with different distributions,
sources, and types \cite{Lv2017NextGeneration}. Compared to traditional big
data, multimodal big data is composed of several modalities to describe the same
thing. For example, an image and text information are often used together to
describe an event in a newspaper. The fusion of information from multimodal data
can be broadly classified into three groups, the early fusion, late fusion, and
intermediate fusion \cite{Gaw2022Multimodal}. The early fusion combines features
from multi-modalities before the neural network training. For example, the
eigenvector can be used as a representative of a data source. Then the
combination of eigenvectors from multimodal data can be used as input of a
traditional classifier, such as a Support Vector Machine(SVM)
\cite{Liu2017Deep}. On contrary to the early fusion, the late fusion combines
information from multimodal data after the training process. For example, after
obtaining the prediction results based on each modality data, the final
prediction can be made by using their averaging values or maximum values
\cite{Ramachandram2017Deep}. Lastly, in order to construct an end-to-end
framework of multimodal learning, the intermediate fusion has been widely
proposed. A typical process of intermediate fusion includes three steps, (1)
each modality is embedded into a latent space using a neural network layer, (2)
the representations of each modality is fused into a single representative, and
(3) a joint representation is learned to make a single prediction by using the
step 2 as inputs \cite{Gaw2022Multimodal}.

\section{Methods}
The following sections introduce the developed GMu-SGCN model and the two types
of its variants, i.e., the Regional Spatially Graph Convolutional Neural Network
(RSGCN) and the Edge Spatially Graph Convolutional Neural Network (ESGCN). In
order to evaluate the performance of the developed model, the GMu-SGCN model and
its variants are used for predicting the edge existence with given node
locations and spatial environment. This section also introduces the developed
framework for edge existence prediction.

\subsection{GMu-SGCN Model}
\begin{figure*}[!htb]
  \centering
  \includegraphics[width=\textwidth]{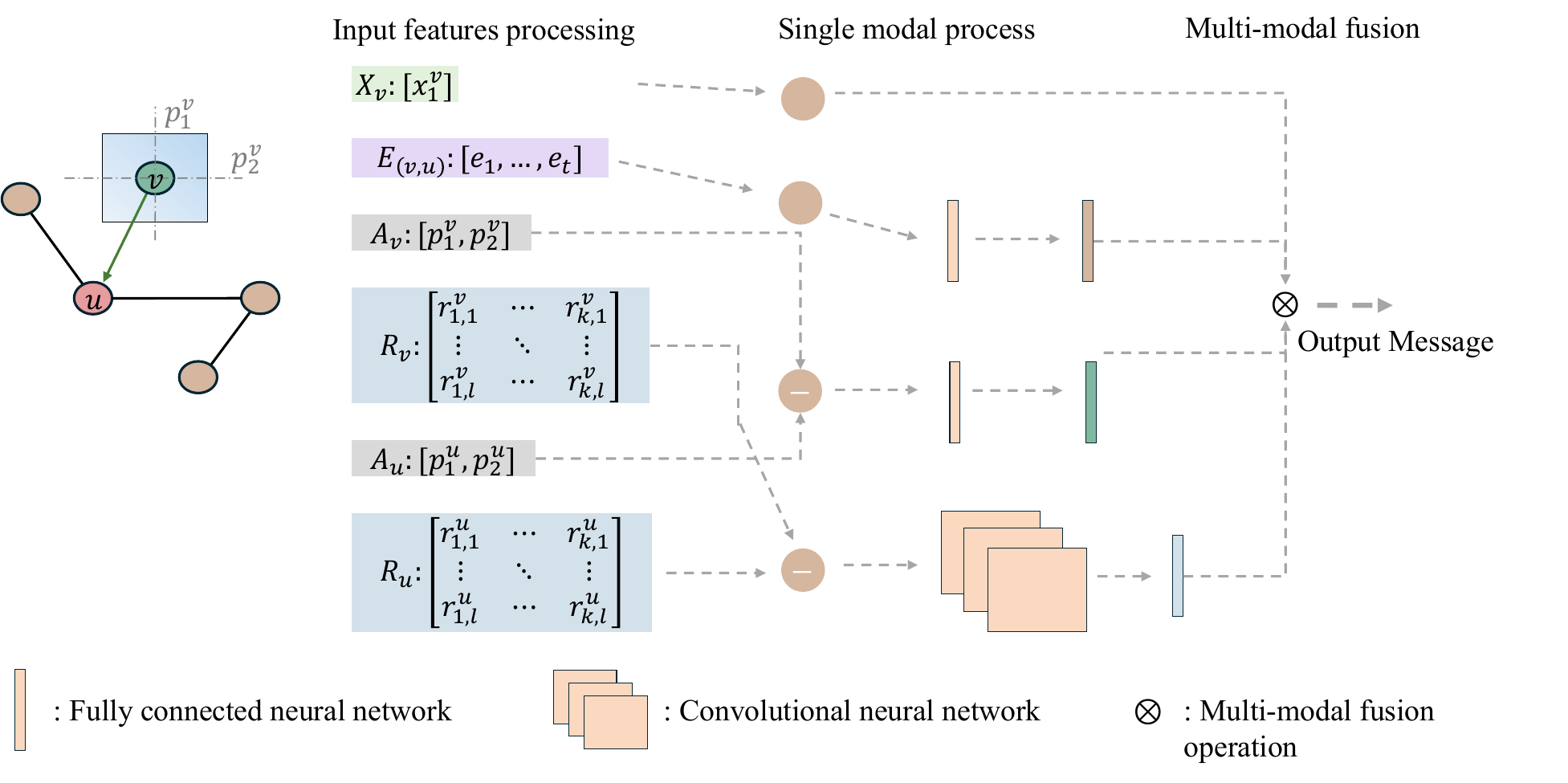}
  \caption{GMu-SGCN layer illustration}
  \label{fig: framework of GMu-SGCN}
\end{figure*}

The developed GMu-SGCN model aims to provide a fusion learning framework for
SENs with the consideration of multimodal features. The framework of the
developed model is shown in Figure \ref*{fig: framework of GMu-SGCN}, which
includes three main components, the input feature manipulation, the single modal
feature processing, and the multimodal feature fusion.

\textbf{Input feature processing:} Processing the input features is a common
approach to improve the neural networks learning efficiency. For example, the
feature normalization and feature selection are often used before the training
process of the machine learning algorithms\cite{Ahmad2019Data,
  Raju2020Study}. For real-world SENs, the relative difference between the
nodes' features often plays a more important role than the absolute feature
values.  For example, the elevation difference between the nodes (grade slope)
is more important than the absolute elevation when designing the road segments
\cite{Hauer2000Safety}. Therefore, in the developed GMu-SGCN model, the node's
point feature, regional feature, and position feature, are firstly normalized
and then the relative difference is calculated when conducting the convolution
process. Specifically, when coevolving the features from a neighbor node $v$ to
a target node $u$, the node's regional and position features of node $v$ are
abstracted by that of the node $u$, as shown in the Figure \ref*{fig: framework
  of GMu-SGCN}. This step aims to improve the learning efficiency of the neural
networks. Experts opinions and domain knowledge can also be incorporated in this
process in the future studies.

\textbf{Single feature embedding:} After conducting the input feature
processing, the processed features are first fed into dedicated neural networks
for a shallow feature extraction. Different types of neural network structures
can be considered based on the data structures. For example, the fully connected
neural networks can be used to extract features from the relative position
feature using Eq \ref{eq: position_feature}. This equation includes the feature
processing described in \textbf{input feature processing}. The edge feature is
extracted by another fully connected neural network (Eq. \ref{eq:
  edge_feature}). Meanwhile, the 2-dimensional convolutional neural networks can
be used to process the relative regional feature (Eq \ref{eq:regional_feature}),
considering the regional feature is a two-dimensional data. In this study, only
a single value from the spatial environment is used as the node's point
feature. Therefore, this feature is directly fed into the next stage without
embedding. For studies with a various number of node's point feature, fully
connected neural networks can also be considered, which is similar to the
process of feature embedding of node's position feature.

\begin{equation}
  \widetilde{p} = \sigma\left[\left(p_u-p_v\right)W_1\right]
  \label{eq: position_feature}
\end{equation}
where $\sigma$ is the \textit{$LeakyReLu$} activation function, $p_u$ is the
position of node $u$, and $W$ is the weights of layers.

\begin{equation}
  \widetilde{p} = \sigma\left[e_{u,v}W_1\right]
  \label{eq: edge_feature}
\end{equation}
where $\sigma$ is the \textit{$LeakyReLu$} activation function, $e_{u,v}$ is the
edge between node $u$ and node $v$, and $W$ is the weights of layers.

\begin{equation}
  \widetilde{r} = \sigma\left(r\cdot K\right)\left(i,j\right)=\sigma\left(\sum_{k}^{m_1}\sum_{l}^{m_2}K_{[k,l]}r_{[i-k,j-l]}\right)  
  \label{eq:regional_feature}
\end{equation}
where $\widetilde{r}$ is the convolved value of the output, $K$ is the kernel
window, and $r$ is the input 2-dimensional regional information, i.e., the
regional spatial data ($a$) in this study. $m_1$ is the height of the input data
and $m_2$ is its width.  $i$, $j$ are the coordinates of the elements in
$\widetilde{r}$.

\textbf{Multimodal fusion and graph convolution:} After each type of feature is
embedded into a shallow representation, the multimodal feature fusion can be
achieved by an element-wise multiply operation, as shown in Eq:
\ref{eq:feature_fusion}. The feature convolution is then conducted by
summarizing of all fused features from the node's neighbors. This convolved
feature replaces the original node's point feature. The convolution process can
be described by Eq: \ref{eq:gcn_process}.
\begin{equation}
  \widetilde{m} = \widetilde{p} \cdot \widetilde{x} \cdot \widetilde{r}
  \label{eq:feature_fusion}
\end{equation}
where $\widetilde{m}$ is the transformed message.

\begin{equation}
  x^l_i = \sigma \left(x^{l-1}_i + \sum_{j \in N_i} \cdot m_j\right)
  \label{eq:gcn_process}
\end{equation}
where $x^l_i$ is the convolved feature of node $i$ at $l^{th}$ layer, $\sigma$
is the \textit{LeakyReLU} activation function, $m_j$ are the transformed
messages from neighbor nodes.
\subsection{The variations of GMu-SGCN}

Given the flexibility in feature embedding and multimodal fusion, the developed
GMu-SGCN model can be modified to specifically working with only node's features
or edge's features. The performance of the model that only considers node
features or edge features has also been compared in this study. For the purposes
of convenience, the variant that only considers node features is named as
Regional Spatial Graph Convolutional Network (RSGCN) and the variant only
considers edge features is named as Edge Spatial Graph Convolutional Network
(ESGCN). Notably, the RSGCN model has been introduced in our previous work
\cite{Fan2024Modelinga}.

\subsection{Link prediction task for SENs}
\label{sec:linkprediction}
In order to evaluate the performance and efficiency of the developed GMu-SGCN
model and its variants, the link prediction task has been used in this
study. Identifying the network connection function is a fundamental challenge in
complex networks and crucial task in real-world applications
\cite{Bhatkar2023Link}. The link prediction task aims to identify the most
efficient model which can learn the network's connection patterns and then
accurately predict the edge existence with given nodes. Both statistical and
deep learning-based methods have been proposed in previous studies
\cite{Dettmann2016Random,Gu2021Learning}.

\begin{figure}[!htb]
  \centering
  \includegraphics[width=\textwidth]{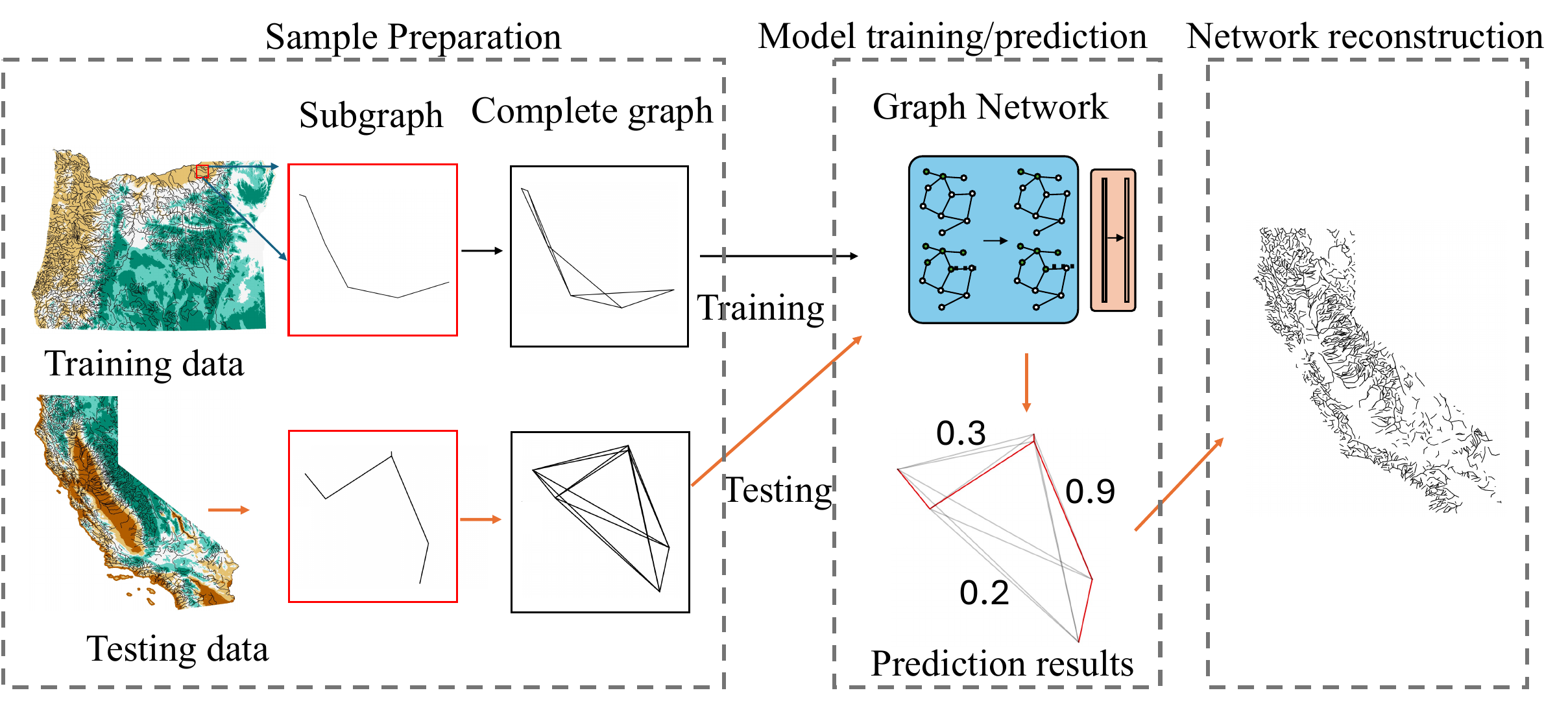}
  \caption{Overall framework for link prediction}
  \label{fig:frameworkofnetworkresemble}
\end{figure}

The overall framework of the link prediction task used in this study is shown in
the Fig \ref{fig:frameworkofnetworkresemble}. The developed framework contains
three major components, i.e., the training samples preparation, model
training/prediction, and network reconstruction. Two types of spatially embedded
networks have been considered in this study, a river network and a power
network. Both river and power network testbeds use data from different states
for model's training and testing. This approach ensures the elimination of the
potential data leakage issue. Details about each component of this framework
have also been introduced in the following sections.

\textbf{Sample preparation:} It is known that the maximum lengths of the
spatially embedded networks are constrained by the spatial environments. A pair
of nodes is unlikely connected when their distances is larger than a
threshold. This is also a common observation in real-world, for example, the
maximum lengths of straight lines in power networks have been statistically
discussed in previous study \cite{Soltan2016Generation}. In this study, to
decrease the computing complexity, a region with a fixed window size is firstly
determined. Then a set of subgraphs is collected by using the sampling window
and each node as the center. The selection of this sampling window size is
dependent on the study area and edge length distribution. A smaller window size
may miss the longer edges, which lead to a incomplete final graph in the network
resembling step. On the other side, a larger window size causes each subgraph
contains too many nodes and edges, which significantly increases the network
complexity and computing memory. In this study, this windows size is selected by
ensuring the smallest subgraphs have at least 3 nodes and 2 edges. The sampled
subgraphs are then simplified by removing the middle nodes and are converted to
complete graphs, as shown in the sample preparation stage of Fig:
\ref*{fig:frameworkofnetworkresemble}. A complete graph is a graph that all
pairs of nodes within the graph are connected. Its edges are labeled as 1 or 0
depending on either this edge exists in the original subgraph or not.

\textbf{Sample training:} The geometric-based deep learning models are then
trained with the prepared samples. The aforementioned geometric-based deep
learning models, i.e., the GMu-SGCN, RSGCN, and ESGCN, are compared with another
geometric deep learning model, GraphSAGE\cite{Hamilton2017Inductive}. The
GraphSAGE is used as a benchmark model in this study as it has been widely used
in various graph learning tasks. The models are firstly used to embed different
types of nodes' and edges' spatial features. Then the connection pattern is
represented by concatenating the features of the corresponding end nodes. As
shown in the Model training/testing stage in
Fig. \ref*{fig:frameworkofnetworkresemble}, a two-layers fully connected neural
network is used for predicting either the edge should be labeled as 1 or 0. The
prediction results of the edge labels vary from 0 to 1, which can be interpreted
as the edge existence probability in this study. More details about the
developed model can be found in the public repository.

\subsection{Model performance evaluation}

The final reconstructed network is resembled by the averaging the prediction
results of all samples. As shown in the last component of
Fig. \ref{fig:frameworkofnetworkresemble}. The network resemble is only
conducted for the test bed dataset.  Given the testing SEN has never been used
in the training process, the performance of different models can be evaluated by
comparing their resembled accuracies. An edge may exist in multiple subgraphs
due to the sampling strategy. In this study, the final edge existence is
determined by using the averaged edge existence probability. An edge is
classified as existence if the averaged existence probability is higher than a
predefined probability threshold. Consequently, this developed resembling
process is highly efficient because this strategy automatically excludes edges
between nodes that are extremely far apart.

\section{Case study}

Two river networks and two power networks are considered in this study.  The
river networks represent the natural developed systems and the power networks
represent man-made infrastructure systems. The following sections provide the
detailed information of both SENs.

\subsection{River Network}

\begin{figure}[!htb]
  \centering
  \subfloat[\centering California River]{{\includegraphics[height=0.4\textwidth]{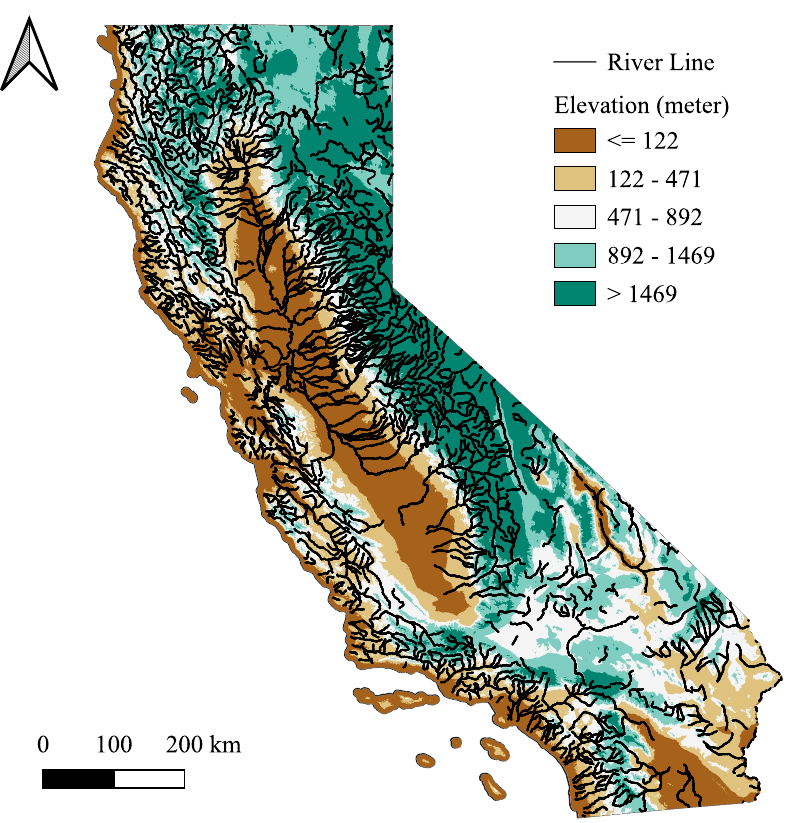} }}%
  \subfloat[\centering Oregon River]{{\includegraphics[height=0.4\textwidth]{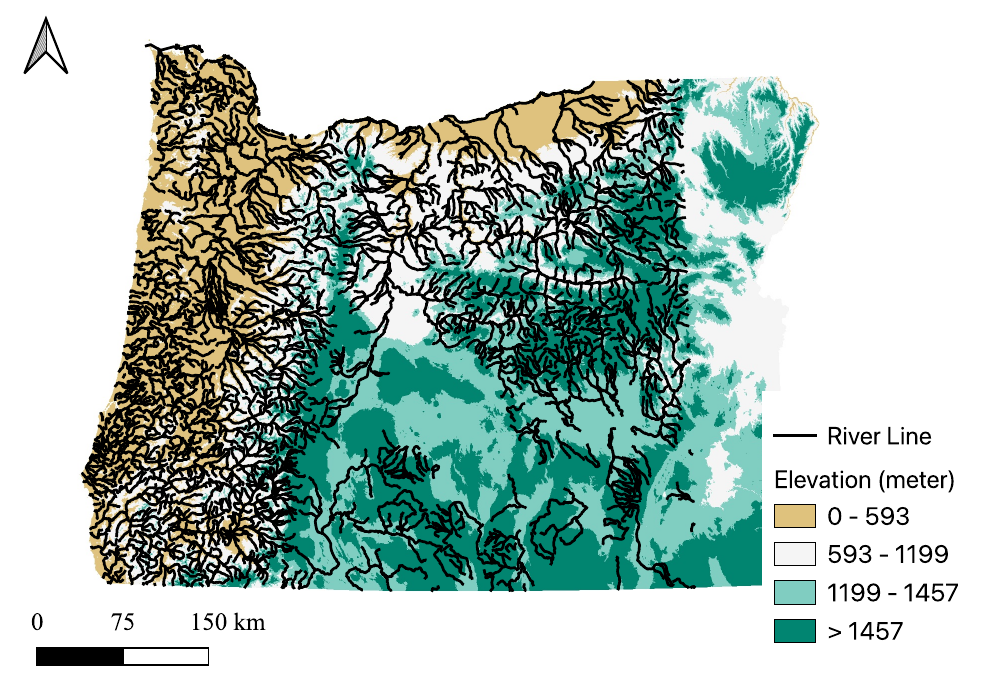} }}%
  \caption{River Network Maps}%
  \label{fig:rivermaps}%
\end{figure}

The river networks located in the Oregon state and California state are selected
for the training and testing purposes, respectively. The states of Oregon state
and California state are contiguous and located in the west side of the
US. Figure \ref{fig:rivermaps} shows the overview of the California river
network and Oregon river network, respectively. The elevation map of both states
have also been visualized. The Oregon river network is used as the training set
and the California river network is used as the testing set. The river networks
are collected from the US National Weather Service
\cite{USDepartmentofCommerceRiversa}. The digital elevation map is downloaded
from the NASA EarthData with a resolution of 30-meters \cite{NASAJPL2013NASA}.

The sampling window size is set at $40km$. Figure \ref{fig:subgraph example
  river} shows an example of the sampled subgraph. Each subgraph contains four
different types of feature that sampled from the digital elevation map, i.e.,
the node's regional feature, the node's point feature, the edge feature, and the
node's position feature. The regional window of each node is set at $1.5km$. The
elevation change within this window is used as the node's regional feature. The
node's elevation value is used as the node's point value. In addition, $128$
elevation values are sampled uniformly along each edge. Table
\ref{tabl:riverstatistics} shows the statistic values of the whole networks and
the sampled subgraphs of river networks in California and Oregon. The table
shows that the river edge numbers in California and Oregon are close to each
other, i.e. $4,392$ and $4,274$ respectively. After sampling the subgraphs from
the original networks using the determined sampling window size, the average
node number of the subgraphs in California river is $10.8$, whereas the average
node number is $13.0$ in Oregon. The number implies that the subgraphs sampled
from Oregon is denser than that from California. The sampled subgraphs are
transformed into complete graphs as mentioned earlier. It can be seen that the
average edges of the complete subgraphs and real subgraphs of Oregon are larger
than that of California, indicating that the subgraphs of Oregon contains more
edges and potential connections.

\begin{figure}[!htp]
  \centering
  \includegraphics[width=0.5\textwidth]{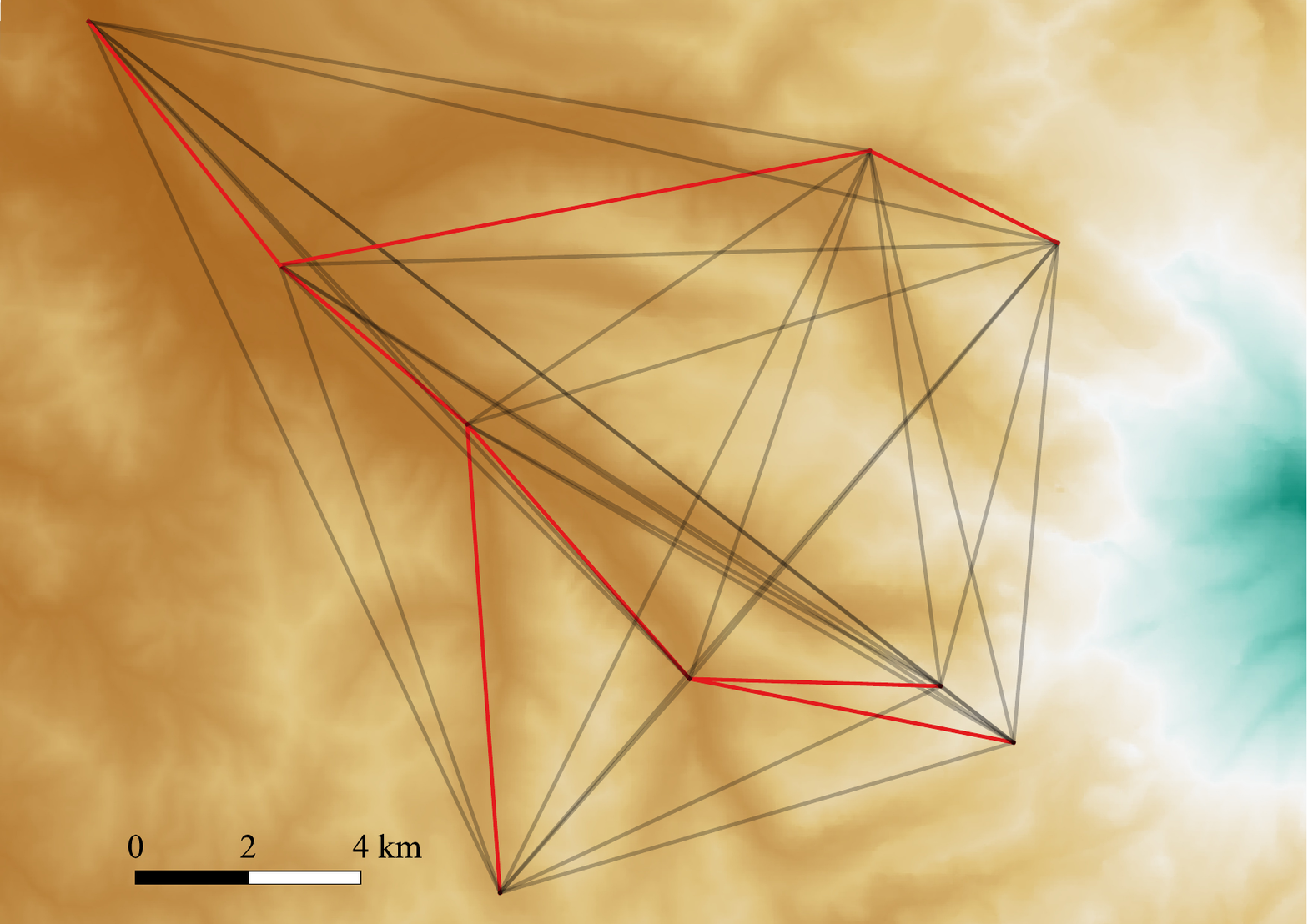}
  \caption{A random subgraph sampled from Oregon River Network}
  \label{fig:subgraph example river}
\end{figure}

\begin{table}[!htp]
  \centering\caption{Statistic comparison of Oregon and California river networks (Size represents the total edge number
    of original network. The node and edge represent the distribution of sampled dataset)}\label{tabl:riverstatistics}
  \begin{tabular}{cc|ccc|ccc|ccc}\toprule
    city & size  & \multicolumn{3}{c}{node} & \multicolumn{3}{c}{edge (complete)} & \multicolumn{3}{c}{edge (real)}                                    \\ \hline
         &       & min                      & mean                                & max                             & min & mean & max &
    min  & mean  & max                                                                                                                                 \\  \hline
    CA   & 4,392 & 6                        & 10.8                                & 44                              & 15  & 66.2 & 946 & 5 & 9.8  & 43 \\ \hline
    OR   & 4,274 & 6                        & 13.0                                & 44                              & 15  & 97.1 & 946 & 5 & 12.1 & 43 \\
    \bottomrule
  \end{tabular}\end{table}

\subsection{Power Network}

\begin{figure}[!htb]
  \centering
  \subfloat[\centering New Jersey Power]{{\includegraphics[height=0.4\textwidth]{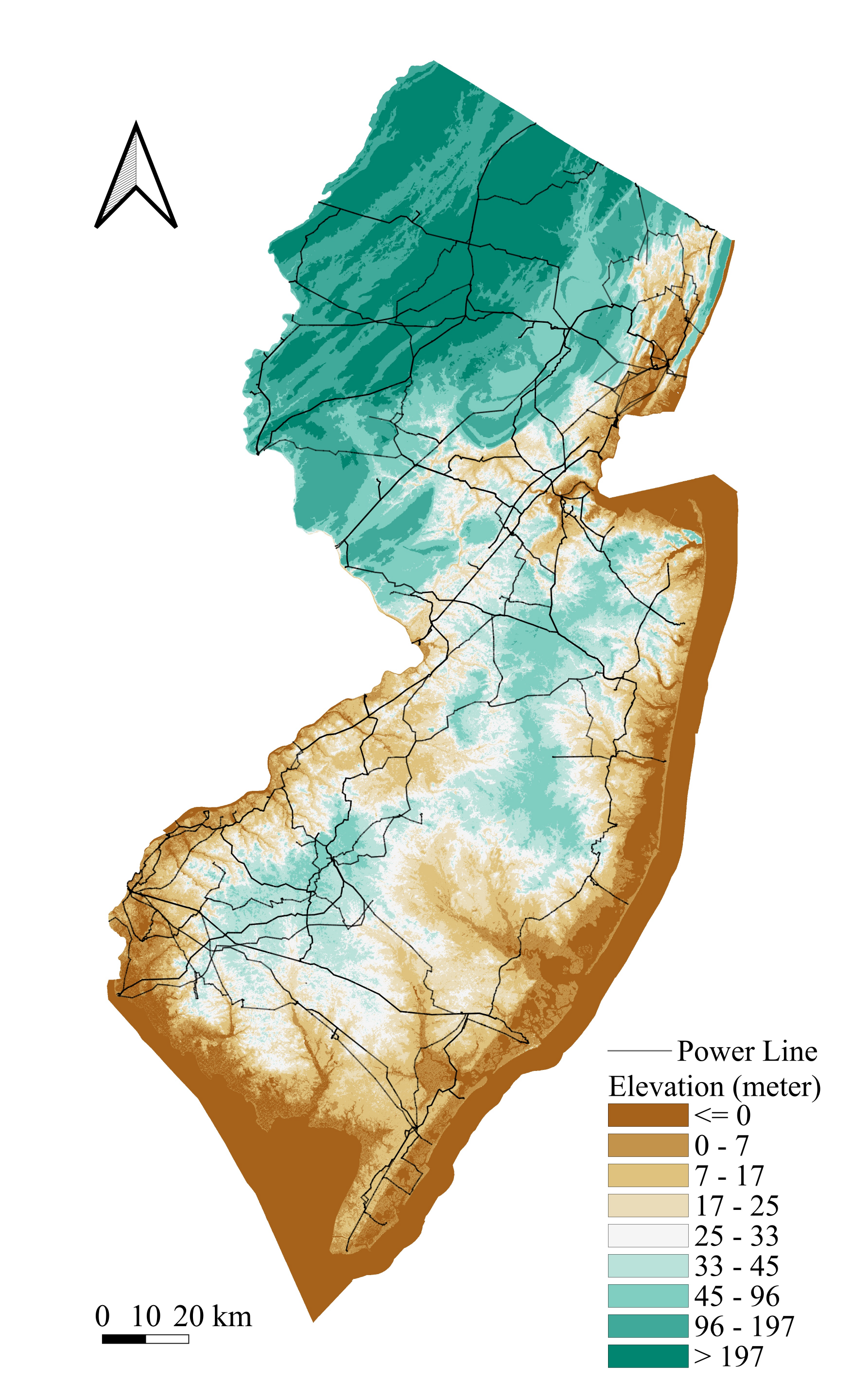} }}%
  \qquad
  \subfloat[\centering Connecticut Power]{{\includegraphics[height=0.4\textwidth]{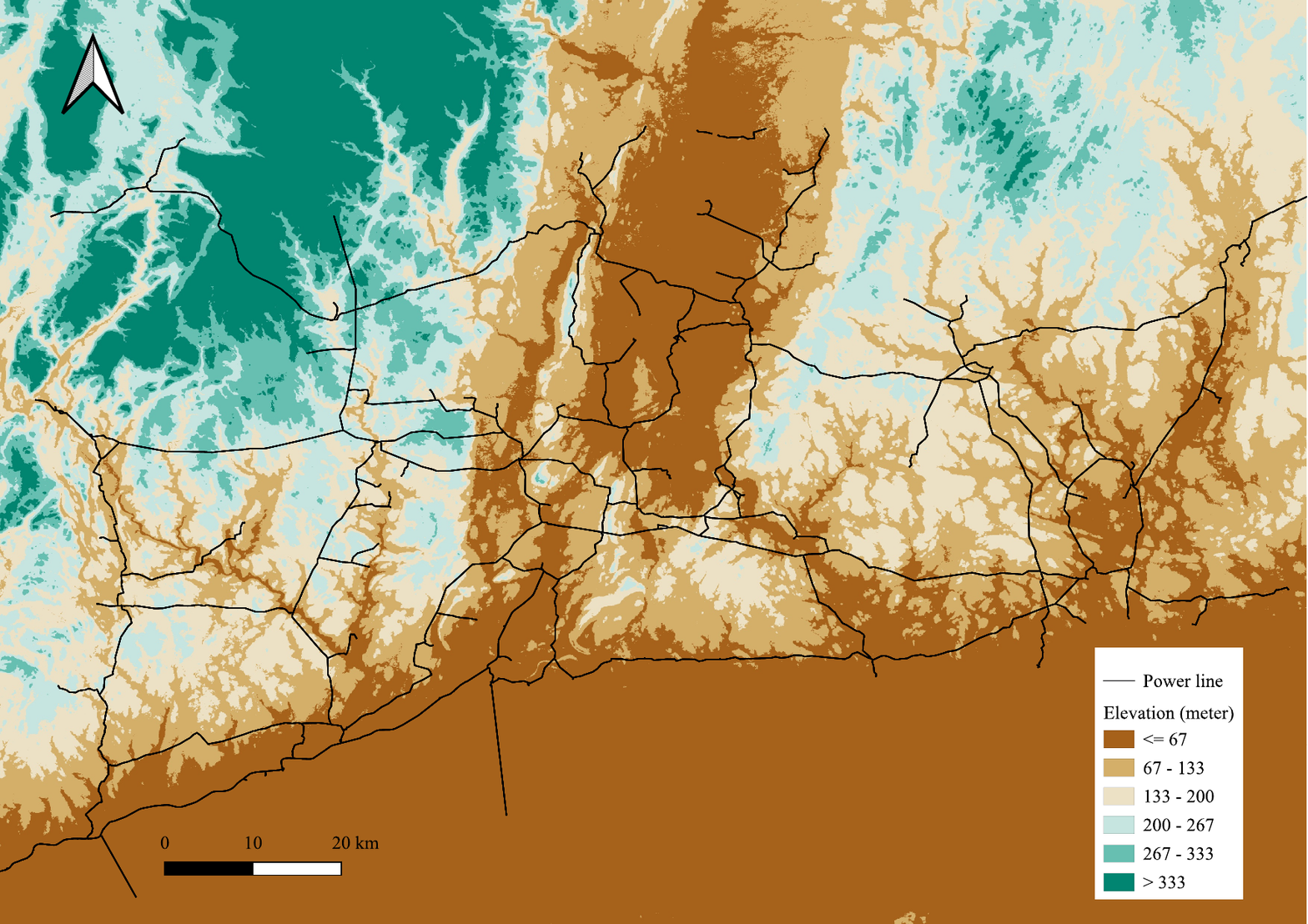}}}
  \caption{Power Network Maps}%
  \label{fig:powermaps}%
\end{figure}

The New Jersey and the Connecticut are two states the located in the eastern
part of the US. The transmission network data from both states is obtained from
the Homeland Infrastructure Foundation-level Data (HIFLD)
\cite{HIFLD2024Homeland}, which contains the national-wide transmission network
varying from 69 kV up to 765 kV. The original graph is cleaned by merging close
and parallel lines into a single line. The sampling window size is
$20km$. Figure \ref{fig:powermaps} shows the overall map of the New Jersey power
network and the Connecticut power networks with the digital elevation
information. The New Jersey network is used as the training data and the
Connecticut network is used as the testing data.

Table \ref{tabl:powerstatistics} also summarizes the node and edge numbers of
the original networks and subgraphs. It can be seen that the power networks used
in NJ is relatively larger than that in Connecticut ($1,609$ vs $633$), mainly
because the area of Connecticut is much smaller than the New Jersey. However,
the sizes of the subgraphs of NJ and CT are relatively similar. For example, the
average node number of the subgraphs in NJ is $22.7$, whereas the number of CT
is $17.2$. The average number of edges of the complete subgraphs and real
subgraphs of NJ are larger than that of CT, indicating the overall subgraphs
sampled from NJ is denser than subgraphs sampled from CT. It should be noted
that although the maximum node number of the NJ dataset is only slightly larger
than that of CT dataset ($57$ vs $34$), the number of potential connections is
significantly higher ($1596$ vs $561$). The main reason is that the potential
connections would quadratically increase with the increase of node numbers. The
comparison also highlights the importance of partitioning a large network into a
set of subgraphs to reduce the computing intensity.

\begin{table}[!htp]
  \centering\caption{Statistic comparison of New Jersey and Connecticut power networks}\label{tabl:powerstatistics}
  \begin{tabular}{cc|ccc|ccc|ccc}\toprule
    city & size  & \multicolumn{3}{c}{node} & \multicolumn{3}{c}{edge (complete)} & \multicolumn{3}{c}{edge (real)}                                       \\ \hline
         &       & min                      & mean                                & max                             & min & mean  & max   &
    min  & mean  & max                                                                                                                                    \\  \hline
    NJ   & 1,609 & 3                        & 22.7                                & 57                              & 3   & 298.4 & 1,596 & 2 & 22.6 & 61 \\ \hline
    CT   & 633   & 4                        & 17.2                                & 34                              & 6   & 162.9 & 561   & 3 & 16.5 & 35 \\
    \bottomrule
  \end{tabular}\end{table}

\section{Results and Discussion}

\subsection{Evaluation}
\label{sec:evaluation}

The considered models are compared by the edge existence prediction
performance. Given the edge existence prediction is essentially a binary
classification problem, the F1-score and accuracy are used for the evaluation
purposes.

Eq. \ref{eq:f1score} shows the definition of the F1-Score, where the True
Positive ($\mathrm{TP}$) represents the edges that are originally existent and
also predicted as existent. The True Negative ($\mathrm{TN}$) represents the
edges that were originally non-existent and also predicted as non-existent. The
False Positive ($\mathrm{FP}$) represents the edges that are originally
non-existent but predicted as existent. And the False Negative ($\mathrm{FN}$)
represents the edges that are originally existent but predicted as
non-existent. The F1-score evaluates the average performance of the model on the
prediction of positive and negative classes.

\begin{equation}
  \mathrm{F1} = \frac{\mathrm{TP}}{\mathrm{TP}+\frac{1}{2}\left(\mathrm{FP}+\mathrm{FN}\right)}
  \label{eq:f1score}
\end{equation}

The existence accuracy represents the percentage of original edges which are
accurately predicted, which can be mathematically represented by
Eq. \ref{eq:accuracy}. The $pred(e_i)$ equals to 1 if the edge is predicted as
'existence', otherwise $pred(e_i)$ equals to 0. Compared to the F1-score metric,
this metric only evaluates the accuracy of the prediction results of the
positive class.

\begin{equation}
  \mathrm{acc} = \frac{\sum_{i=1}^n{pred(e_i)}}{n}
  \label{eq:accuracy}
\end{equation}
where $n$ is the number of total edges in original SEN.

\subsection{Prediction results of the River Network}

\begin{figure}[!htb]
  \centering
  \subfloat[\centering GMu-SGCN]{{\includegraphics[width=0.4\textwidth]{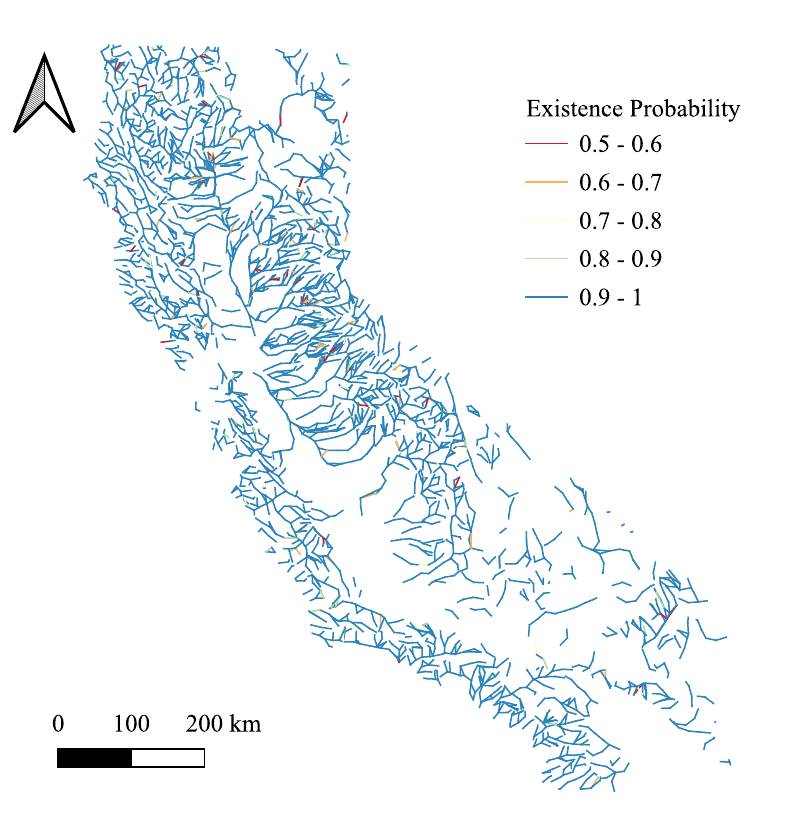}}}%
  \subfloat[\centering RSGCN]{{\includegraphics[width=0.4\textwidth]{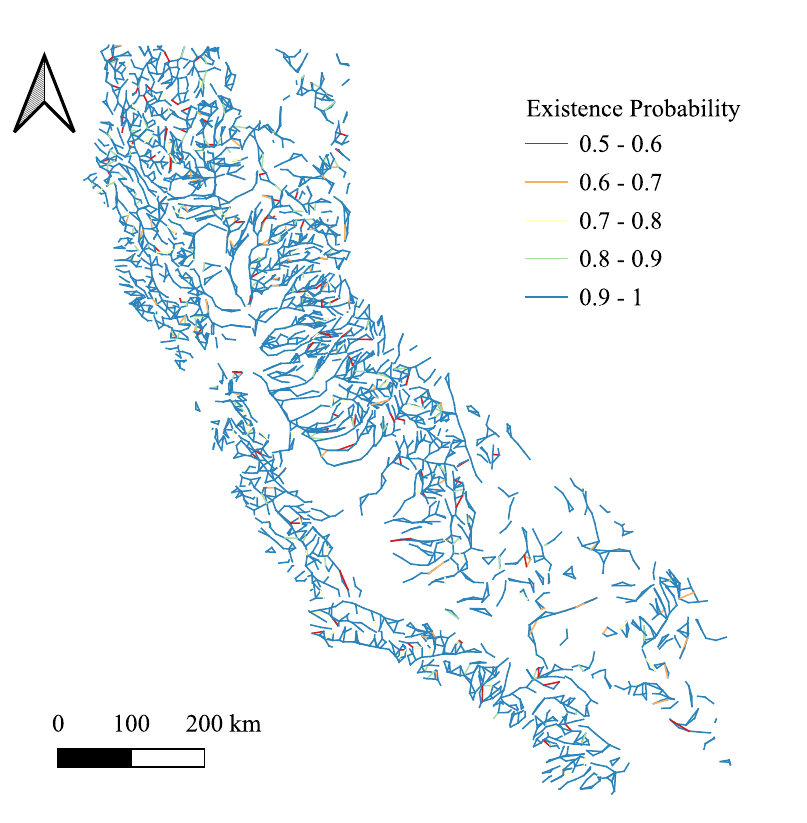}}}%
  \\
  \subfloat[\centering ESGCN]{{{\includegraphics[width=0.4\textwidth]{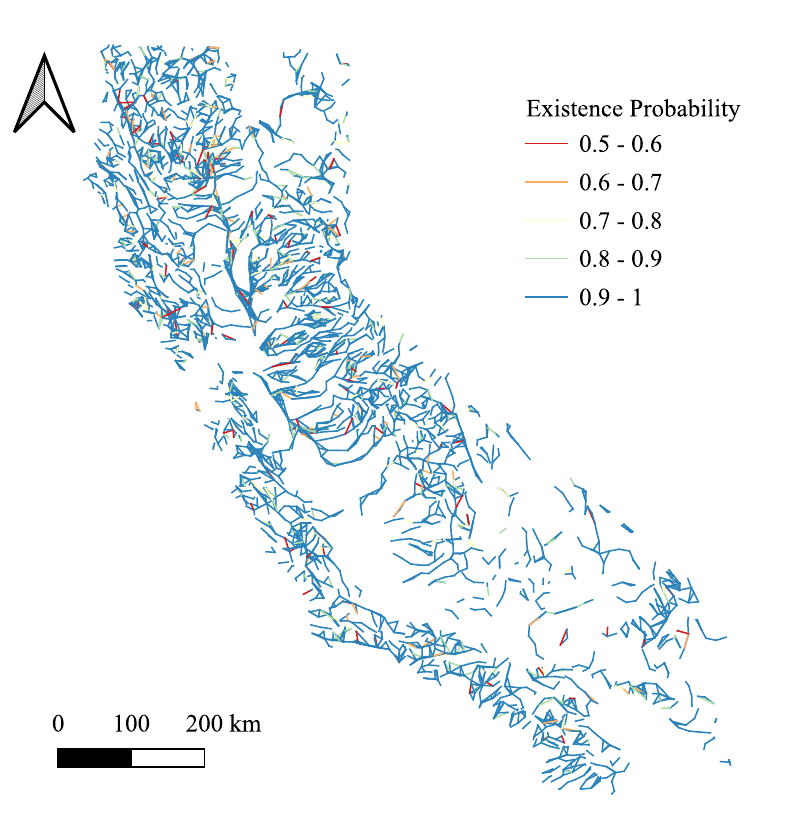}}}}%
  \subfloat[\centering GraphSAGE]{{\includegraphics[width=0.4\textwidth]{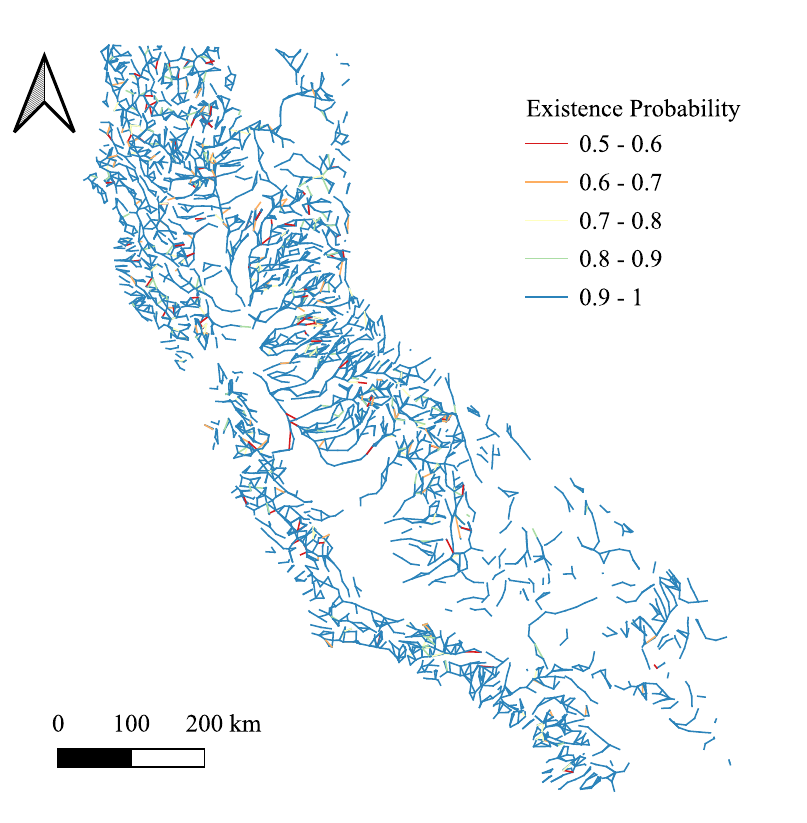} }}%
  \caption{Predicted River Networks}%
  \label{fig:riverresults}%
\end{figure}

Fig \ref{fig:riverresults} shows the reconstruction results of the river network
located in California by the considered models. Only edges whose predicted
existent probability higher than 0.5 are plotted. It can be seen that all
considered models have certain level of capability to reconstruct the river
network. In addition, most edges are predicted with a higher probability (larger
than 90 \%) by all considered models. Only a few of the edges are predicted with
a probability lower than 60 \%. Fig: \ref{fig:riverresults} also shows the
GMu-SGCN model has the lowest uncertain edges compared to other models. Only a
few edges are predicted with a probability that lower than 90\%. In addition,
the predicted network has fewer false positive edges, compared to that of RSGCN,
ESGCN and GraphSAGE.  The network reconstructed by the GMu-SGCN model is more
similar to the original network as shown in Figure: \ref{fig:rivermaps} (a).

\begin{figure}[!htb]
  \centering
  \subfloat[\centering F1-score]{{\includegraphics[width=0.45\textwidth]{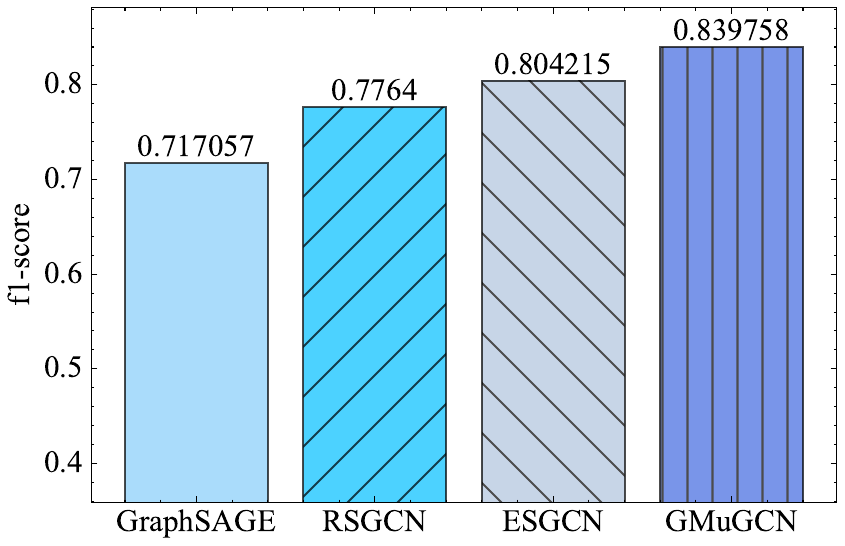} }}%
  \subfloat[\centering Accuracy]{{\includegraphics[width=0.45\textwidth]{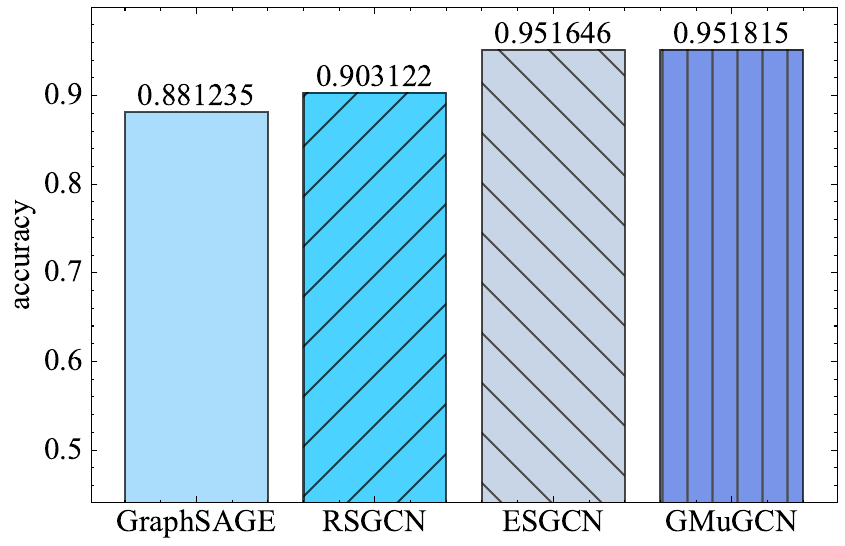} }}%
  \caption{Evaluation metrics for river network}%
  \label{fig:scoreRiver}%
\end{figure}

The models have also been evaluated by using the F1-score and accuracy as
discussed in section \ref{sec:evaluation}. The F1-score evaluates the accuracy
of the prediction of both exist edges and nonsexist edges in the sample dataset,
whereas the accuracy only evaluates the prediction accuracy of original exist
edges. The evaluation metrics are shown in Figure \ref{fig:scoreRiver}. As can
be seen, the developed GMu-SGCN achieved the highest performance in both
F1-score and accuracy metrics. In particular, it is 12.3\%higher than the worst
model, GraphSAGE, and 3.67 \% higher than the second-best model, ESGCN in the
F1-scores. The results of the F1-scores indicate the GMu-SGCN achieved the best
balance in avoiding the True Negative errors and False Positive errors.  On the
other hand, the accuracy of the GMu-SGCN model is very similar to that of the
ESGCN model. The results indicate that the GMu-SGCN and ESGCN have similar
performance in predicting the original exist edges. It can be inferred that the
ESGCN predicted more nonsexist edges as exist edges by comparing Figure
\ref{fig:scoreRiver} (a) and (b). The results also indicate that the connection
patterns of the river network is more dependent on the edge features.

\subsection{Reconstruction results of the Power Network}

\begin{figure}[!htb]
  \centering
  \subfloat[\centering Sampled edges]{{\includegraphics[width=0.45\textwidth]{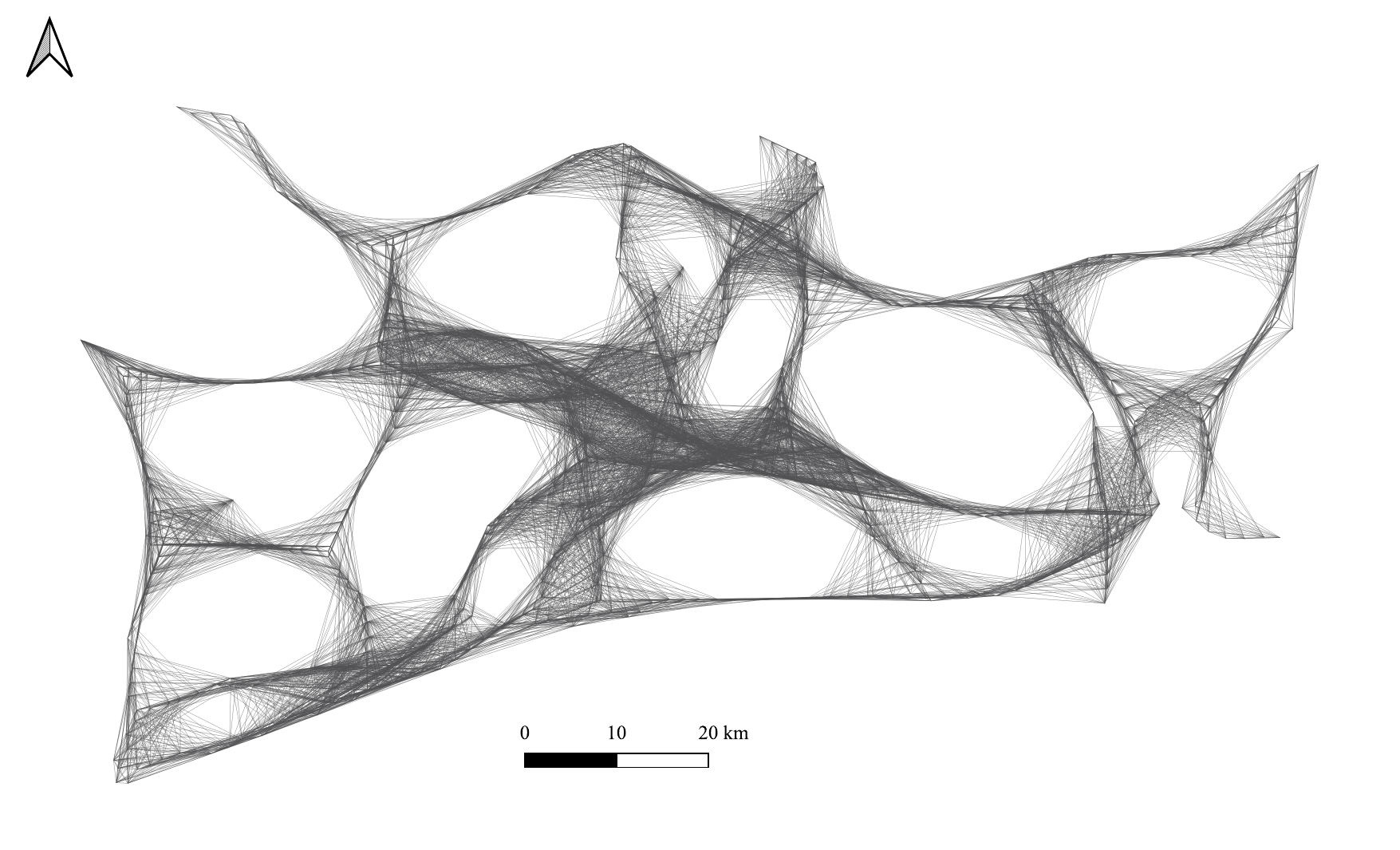} }}%
  \subfloat[\centering GMuSGCN]{{\includegraphics[width=0.45\textwidth]{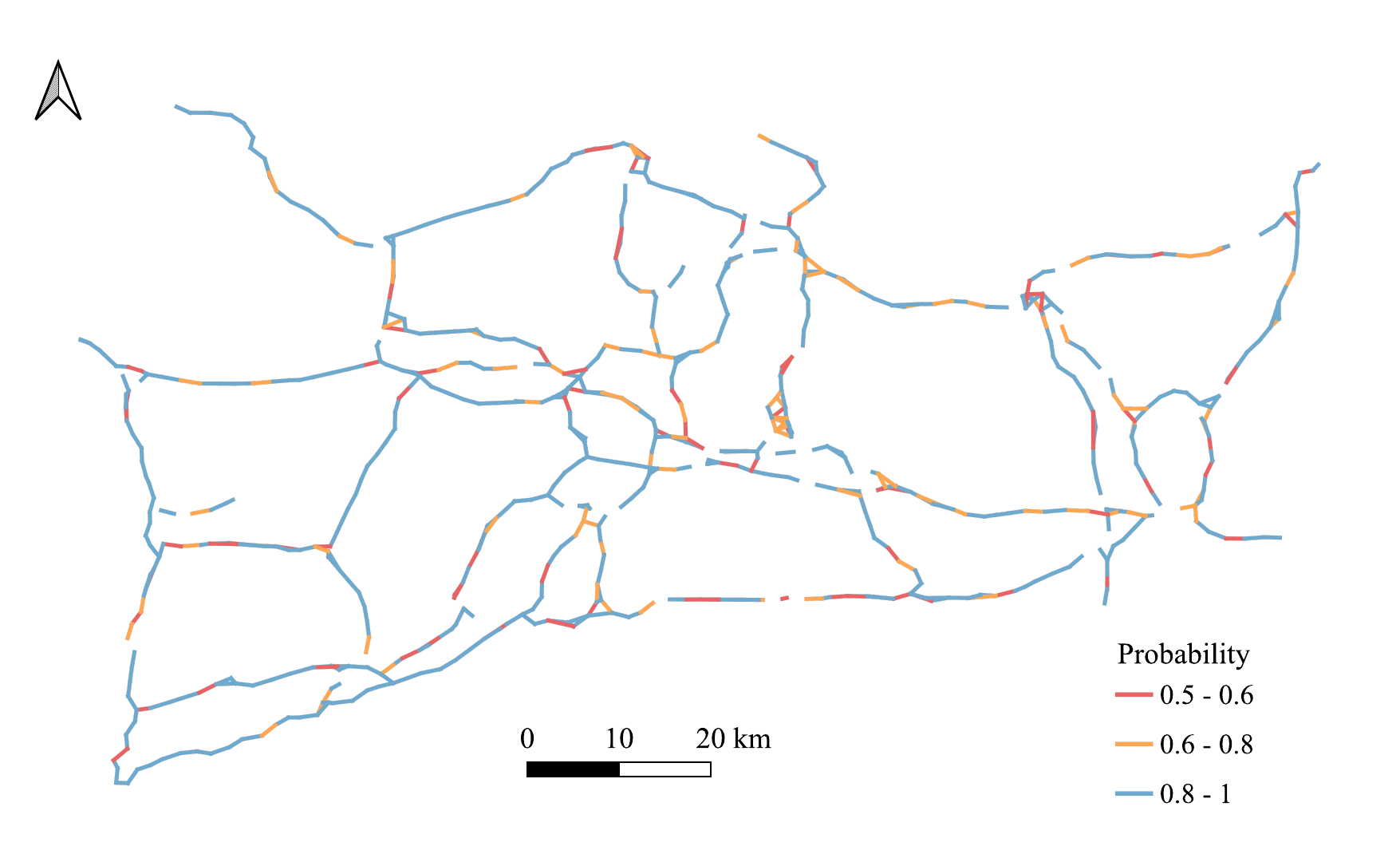} }}%
  \caption{Example of the samples and predicted results by GMuSGCN}%
  \label{fig:powerresults}%
\end{figure}

As aforementioned, the NJ power network is used as the training set and the CT
power network is used as the testing set.  Figure \ref*{fig:powerresults} (a)
shows all potential edges of CT power network based on the sampling method, as
described in section \ref{sec:linkprediction}. It can be seen that only nodes
within the distance of sampling window size are potentially connected. The idea
of this sampling aligns with many observed connection patterns in spatially
embedded networks. Figure \ref*{fig:powerresults} (b) shows the final
reconstruction results by GMu-SGCN model. Only the edges whose predicted
existence probability higher than 0.5 are visualized. Similar to the river
network, most of the edges are predicted with a higher confidence (higher than80
\%). Only a few edges are predicted with a relative low confidence (50\% to
60\%).

The performance of all the considered models is also summarized in Fig
\ref{fig:powerresults}. It can be found that the developed GMu-SGCN model
outperforms the other models. For example, the f1-score of the GMu-SGCN model is
37.1 \% higher than the worst performance model, ESGCN. It is 1.13\% higher than
the second-best model, the RSGCN model. The accuracy of the developed GMu-SGCN
model is 19.1\% higher than the ESGCN model and 3.9\% higher than the RSGCN
model.

\begin{figure}[!htb]
  \centering
  \subfloat[\centering F1-score]{{\includegraphics[width=0.45\textwidth]{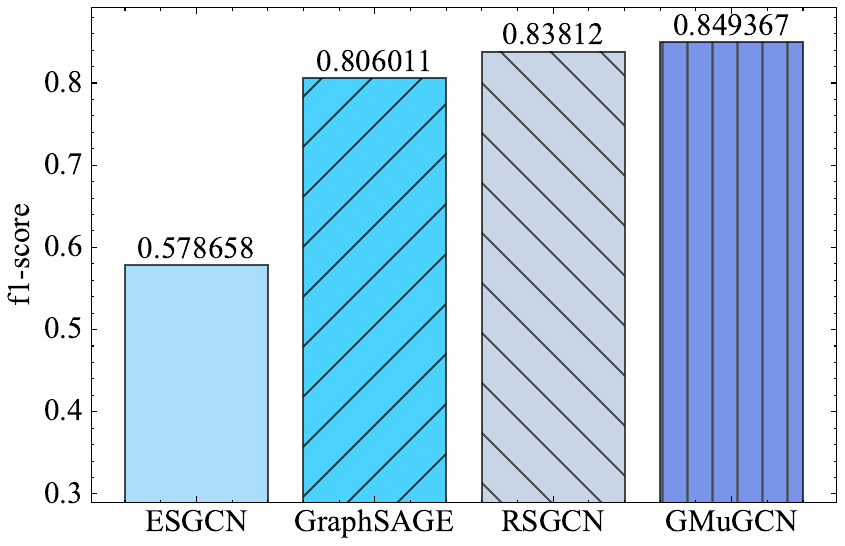} }}%
  \subfloat[\centering Accuracy]{{\includegraphics[width=0.45\textwidth]{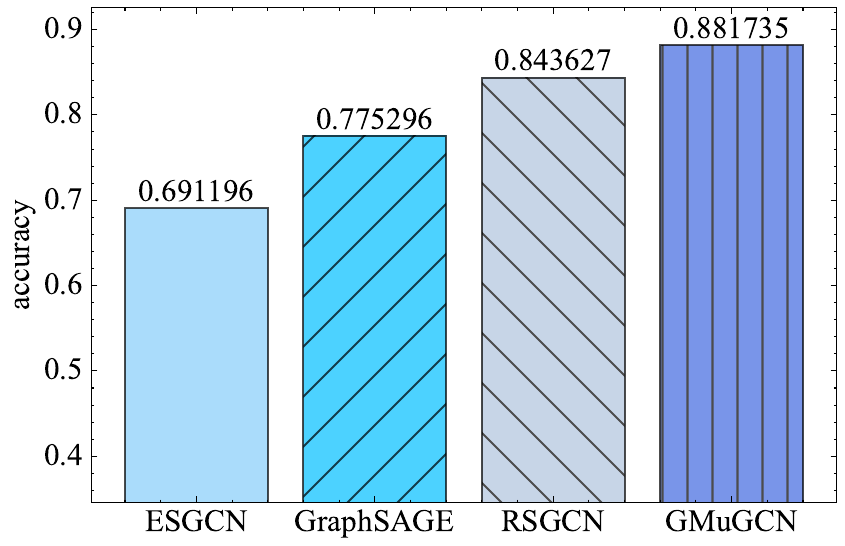} }}%
  \caption{Evaluation metrics for power network}%
  \label{fig:scorepower}%
\end{figure}

The results also show that different SENs are influenced differently by the
embedded environments differently. For example, it can be seen that the ESGCN
model is the second-best model for the river network testbed, indicating the
model is more suitable for SENs who are more sensitive to the edge features. On
the contrary, the RSGCN model outperformed the ESGCN model in the power network
case study, indicating the power network is more sensitive to the node's
regional feature compared to the edge's feature.  Using the ESGCN model and
RSGCN model may benefit a lower memory consumption and higher computing
efficiency compared to directly use GMu-SGCN model. However, the GMu-SGCN model
outperformed all the other models in both test beds.

\section{Conclusion}
In this study, a generic multimodal graph convolutional neural network is
developed for efficient network representation learning. Given the flexibility
of the developed GMu-SGCN model, two variants have also been designed, i.e., the
RSGCN model and ESGCN model. The former model only embeds the node's multimodal
features, whereas the later model only considers the node's edge features. The
network connection prediction task was conducted to evaluate the models'
performance.  Specifically, each model was used to embed various node and edge
features into latent vectors of nodes, and then the edge existence probability
is predicted by using these latent vectors. Two real-world spatially embedded
networks, the river networks and power networks have been used as the test
beds. The results show that the developed GMu-SGCN model outperformed the other
models in all test beds. Specifically, the GMu-SGCN model outperformed the
GraphSAGE model, a widely used GCN model, 12.3\% in river network test bed and
37.1\% in the power network test bed. Furthermore, the RSGCN variant and ESGCN
variant are the second-best model for river network and power network,
respectively. This result indicates that the connection of river network relies
more on their edge features, whereas the connection of power network relies more
on the node features.

Although the developed models can efficiently learn the representation of the
considered SENs, there are some limitations which should be considered in the
future studies. Firstly, the existence of all edges in the sample graphs are
predicted simultaneously. Although this approach is computational efficient, it
does not predict the edge existence in a sequential approach as traditional
methods. As a result, this approach cannot leverage the connection patterns of
previous established edges, and it also cannot guarantee all nodes within a
graph are connected. However, it should be noted that this approach can reduce
the accumulated errors that existed in previous methods. The second limitation
is that the developed method assumes the nodes' positions are known. However,
such information is often missing in many real-world applications. Further
studies should integrate the prediction of nodes into the developed framework.

\backmatter

\bmhead{Acknowledgements}

This work has been supported by a grant from the Energy Research Fund
administered by the Andlinger Center for Energy and the Environment as well as
the School of Engineering and Applied Science (SEAS) Seed Grant at Princeton
University. Furthermore, the authors are pleased to acknowledge that the work
reported on in this paper was substantially performed using the Princeton
Research Computing resources at Princeton University which is a consortium of
groups led by the Princeton Institute for Computational Science and Engineering
(PICSciE) and Office of Information Technology's Research Computing.

\bibliography{manuscript.bib}

\end{document}